%% file: nips_2018.tex
\documentclass{article}

\PassOptionsToPackage{numbers, compress}{natbib}

\usepackage[preprint]{nips_2018}

\usepackage[utf8]{inputenc} 
\usepackage[T1]{fontenc}    
\usepackage{hyperref}       
\usepackage{url}            
\usepackage{booktabs}       
\usepackage{amsfonts}       
\usepackage{nicefrac}       
\usepackage{microtype}      
\usepackage{amsmath}
\usepackage{extpfeil}
\usepackage{algorithm}
\usepackage{algorithmic}
\usepackage{bm}
\usepackage{graphicx}
\usepackage{subfigure}

\title{DEEPEYE: A Compact and Accurate\\ Video Comprehension at Terminal Devices\\ Compressed with Quantization and Tensorization}

\author{
  Yuan Cheng\\
  Shanghai Jiao Tong University\\
  \texttt{cyuan328@sjtu.edu.cn} \\
  \And
  Guangya Li \\
  South University of Science and Technology\\
  \texttt{11749189@mail.sustc.edu.cn} \\
  \And
  Hai-Bao Chen \\
  Shanghai Jiao Tong University\\
  \texttt{haibaochen@sjtu.edu.cn} \\
  \And
  Sheldon X.-D. Tan\\
  University of California, Riverside\\
  \texttt{stan@ece.ucr.edu} \\
  \And
  Hao Yu \\
  South University of Science and Technology\\
  \texttt{yuh3@sustc.edu.cn} \\
}

\begin{document}

\maketitle

\begin{abstract}

As it requires a huge number of parameters when exposed to high dimensional inputs in video detection and classification, there is a grand challenge to develop a compact yet accurate video comprehension at terminal devices. Current works focus on optimizations of video detection and classification in a separated fashion. In this paper, we introduce a video comprehension (object detection and action recognition) system for terminal devices, namely DEEPEYE. Based on You Only Look Once (YOLO), we have developed an 8-bit quantization method when training YOLO; and also developed a tensorized-compression method of Recurrent Neural Network (RNN) composed of features extracted from YOLO.
The developed quantization and tensorization can significantly compress the original network model yet with maintained accuracy. Using the challenging video datasets: MOMENTS and UCF11 as benchmarks, the results show that the proposed DEEPEYE achieves $3.994 \times$  model compression rate with only $0.47\%$ mAP decreased; and $15,047 \times$ parameter reduction and $2.87 \times$ speed-up with $16.58\%$ accuracy improvement.

\end{abstract}

\input{secs/introduction}
\input{secs/yolo}

\input{secs/ttrnn}

\input{secs/overall}
\input{secs/result}

\section{Conclusion}
\label{sec:conclusion}

In this paper, we have proposed a compact yet accurate video comprehension framework for object detection and action recognition, called DEEPEYE. It is a RNN network with features extracted from YOLO. The Q-YOLO with an 8-bit quantization and T-RNN with a tensorized-compression are both developed, which can remarkably compress the original network model yet with maintained accuracy. We have tested DEEPEYE on MOMENTS and UCF11 benchmarks. The results show that DEEPEYE can achieve $3.994 \times$ compression with only $0.47\%$ mAP decreased; and $15,047 \times$ parameter reduction and $2.87 \times$ speed-up with $16.58\%$ accuracy improvement. The proposed DEEPEYE can be further implemented at terminal devices towards real-time video analysis.

\bibliographystyle{IEEEtran}
\bibliography{nips2018}

\end{document}

%% file: secs/introduction.tex
\section{Introduction}
\label{sec:introduction}

The success of convolutional neural network (CNN) has resulted in a potential general feature extraction engine for various computer vision applications \cite{lecun1998gradient, krizhevsky2012imagenet}. However, applications such as Advanced Driver Assistance System (ADAS) require a real-time processing capability at terminal devices. Network model compression is thereby quite essential to produce a simplified model with consideration of both compactness and accuracy.

For example, a YOLOv3 \cite{redmon2018yolov3} network contains almost $100$ convolution layers, which dominate the network complexity. As most convolution filter now is a small sized ($3 \times 3$, $5 \times 5$ etc.) operator, network pruning \cite{guo2017software} may not be suited for this type of network. Direct quantization \cite{hashemi2017understanding} however needs additional training to maintain the accuracy. The application of quantization (such as binary) during training \cite{courbariaux2015binaryconnect, liu2018squeezedtext} has shown the promising deep learning network implication with significant network reduction yet maintained accuracy. But there is no reported work to apply trained quantization method to the large-scale network such as YOLO with good accuracy.

Moreover, YOLO \cite{redmon2016you, redmon2017yolo9000} is originally designed for object detection from images. It is unknown how to extend it into video data analysis such as object detection and action recognition. Recurrent Neural Network (RNN) can be applied for sequence-to-sequence modeling with great achievements by exploiting RNN to video data \cite{yao2015describing, ebrahimi2015recurrent, venugopalan2015sequence}. However, the high-dimensional inputs of video data, which make the weight matrix mapping from the input to the hidden layer extremely large, hinders RNN's application. Recent works \cite{ng2015beyond, fernando2016learning, sharma2015action} utilize CNN to pre-process all video frames, which might suffer from suboptimal weight parameters by not being trained end-to-end. Other works \cite{srivastava2015unsupervised, donahue2015long} try to reduce the sequence length of RNN, which neglects the capability of RNN to handle sequences of variable lengths. As such it cannot scale for larger and more realistic video data.
The approach in \cite{yang2017tensor, tjandra2018tensor} compresses RNN with tensorization using the original frame inputs, which has resulted in limited accuracy as well as scalability.

\begin{figure}[htb]
  \centering
  \includegraphics[width=0.8\textwidth]{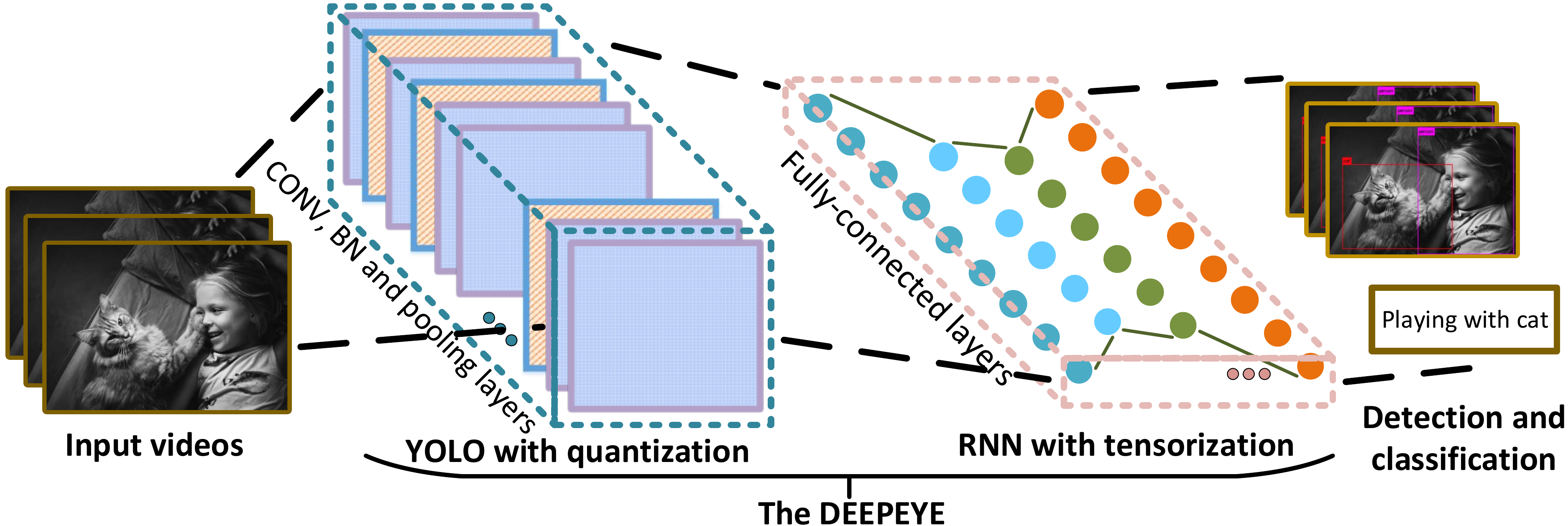}
  \caption{DEEPEYE: a video comprehension framework for object detection and action recognition.}
  \label{fig:deepeye}
\end{figure}

In this paper, we have developed a RNN framework using the features extracted from YOLO to analyse video data. Towards applications on terminal devices, we have further developed an 8-bit quantization of YOLO as well as a tensorized-compression of the RNN. The developed quantization and tensorization can significantly compress the original network model yet with maintained accuracy. Moreover, the above two
optimized networks are integrated into one video comprehension system, which is shown in Fig.\ref{fig:deepeye}. Experimental results on several benchmarks show that the proposed framework, called DEEPEYE, can achieve $3.994 \times$ compression with only $0.47\%$ mAP decreased; and $15,047 \times$ parameter reduction and $2.87 \times$ speed-up with $16.58\%$ accuracy improvement.

The rest of the paper is organized as follows. In Section \ref{sec:yolo} we introduce the basics of YOLO and the YOLO with quantization for real-time video object detection. In Section \ref{sec:ttrnn} we first introduce the tensor-decomposition model and then provide a detailed derivation of our proposed tensorized RNN. In Section \ref{sec:deepeye} we integrate the quantized YOLO with the tensorized RNN as a new framework for video comprehension system, called DEEPEYE. In Section \ref{sec:result} we present our experimental results on several large scale video datasets. Finally, Section \ref{sec:conclusion} serves as a summary of our current contribution and also provides an outlook of future work.

%% file: secs/yolo.tex
\section{YOLO with Quantization}
\label{sec:yolo}

The proposed video object detection structure is based on YOLO, which is a system of frame object detection and
is proposed by using a single convolutional neural network to predict the probabilities of several classes.
In this section, we firstly introduce the basics of YOLO, and then we apply it with 8-bit quantization to maintain a real-time and high-compressed video object detection structure which provides a promising performance on both efficiency and quantity.

\subsection{Basics of YOLO}
\label{sec:yolobasic}

YOLO reframes object detection as a signal regression problem, straight from image pixels of every frames to bounding box coordinates and class probabilities.
A convolutional network simultaneously predicts multiple bounding boxes and class probabilities for those boxes.
YOLO has several benefits over traditional methods of object detection since it trains on full images and directly optimizes detection performance \cite{redmon2017yolo9000}.

\begin{figure}[htb]
  \centering
  \includegraphics[width=0.95\textwidth]{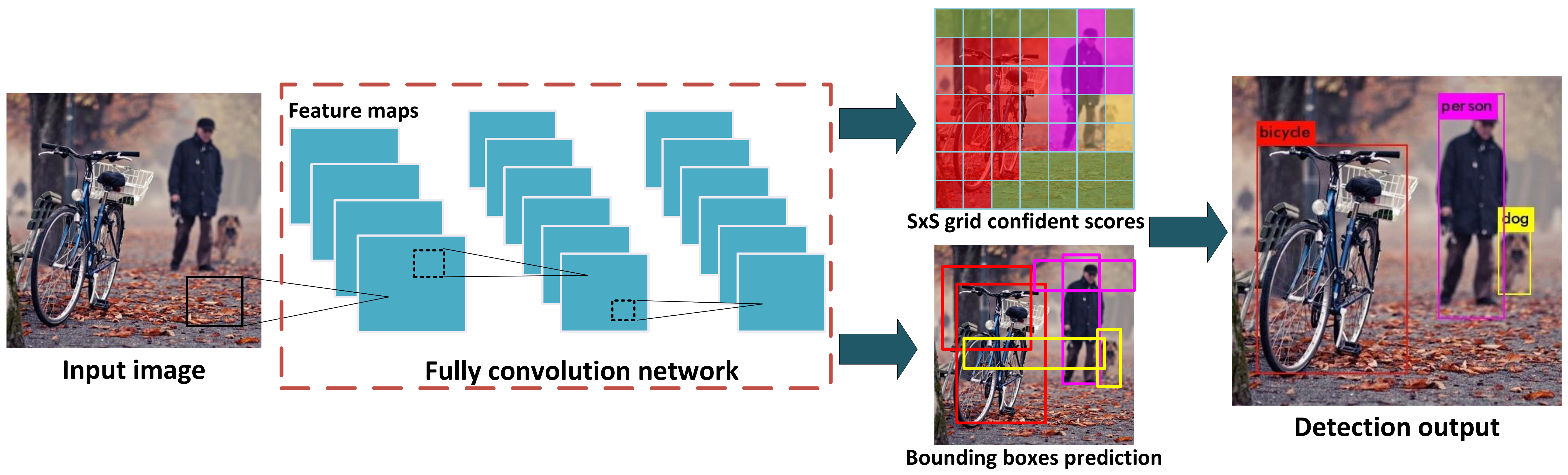}
  \caption{YOLO: object detection of single frame picture.}
  \label{fig:yolo_arch}
\end{figure}

As shown in Fig.\ref{fig:yolo_arch}, it consists of the feature exaction layers and the localization and classification layers, and based on a fully convolution network (FCN) structure. Our system adopts the method to divide the input image into a $S \times S$ grid \cite{redmon2016you}. Every grid cell must be detected if there is an object, then the $B$ (number of boxes) bounding boxes prediction and confidence scores are maintained by the proposed FCN, which will be quantized in Section \ref{sec:quantization}. Confidence is defined as $P_r(Object) \times IOU_{pred}^{truth}$ \cite{redmon2016you, redmon2018yolov3}, which reflects how confident the bounding box contains an object. Here, the intersection over union (IOU) is calculated by using the predicted mask and the ground truth.
For evaluating YOLO on VOC \cite{everingham2010pascal}, if we set the parameters $S = 7$, $B = 2$, then the feature output of the final convolutional layer turns out to be a $7 \times 7 \times 50$ tensor.


\subsection{8-bit-quantized YOLO}
\label{sec:quantization}

The direct YOLO implementation for video-scale data would require large and unnecessary resource of both software and hardware. Previous works in \cite{zhou2016dorefa, zhu2016trained, hubara2016quantized}
suggest a neural network using quantized constraints during the
training process. In this section, we discuss how to generate a
YOLO model (namely Q-YOLO) with 8-bit quantization.


\begin{figure}[htb]
	\centering
	\subfigure[]
	{
		\includegraphics[width=0.35\textwidth]{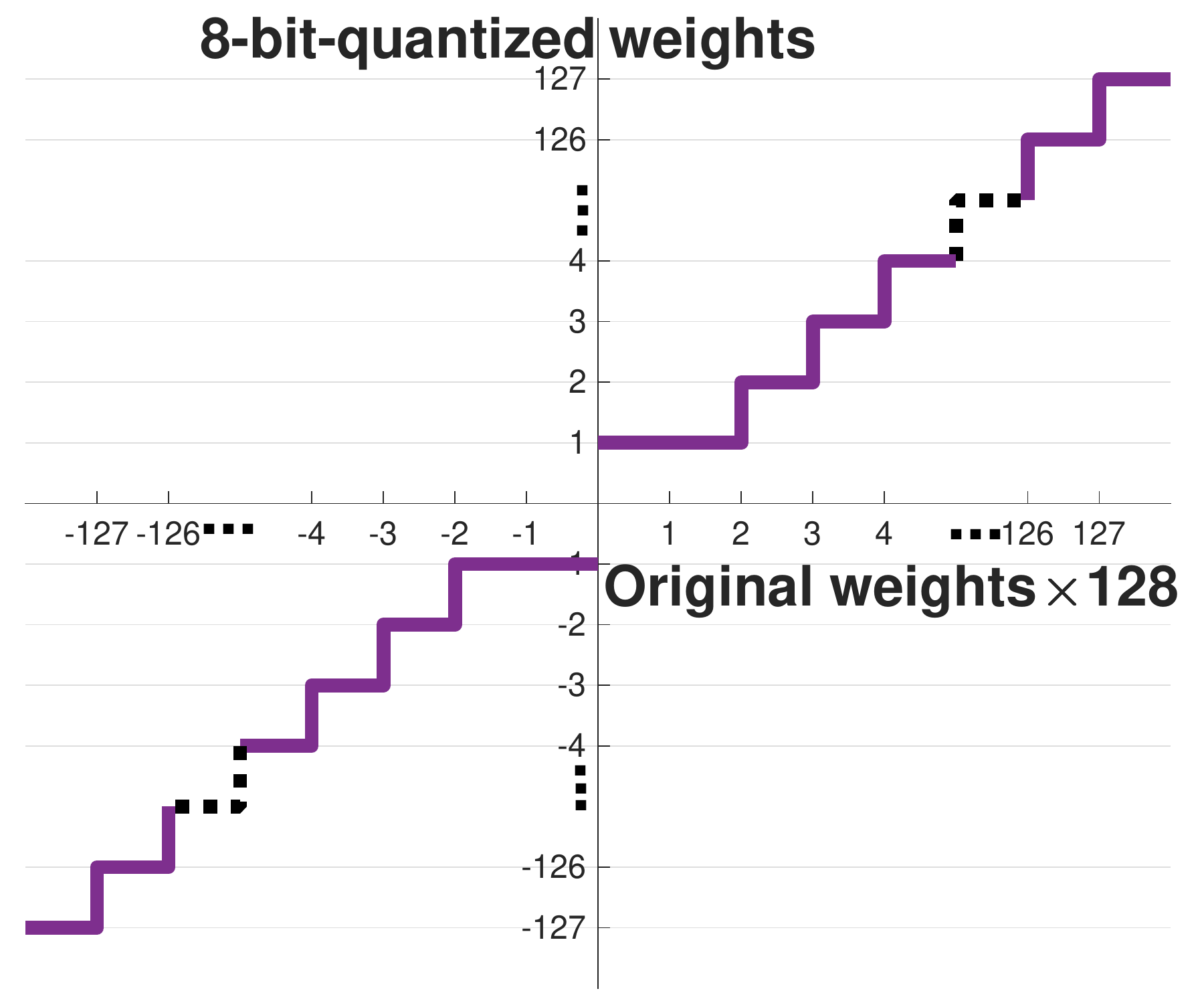}
		\label{fig:weight}
	}
	\subfigure[]
	{
		\includegraphics[width=0.35\textwidth]{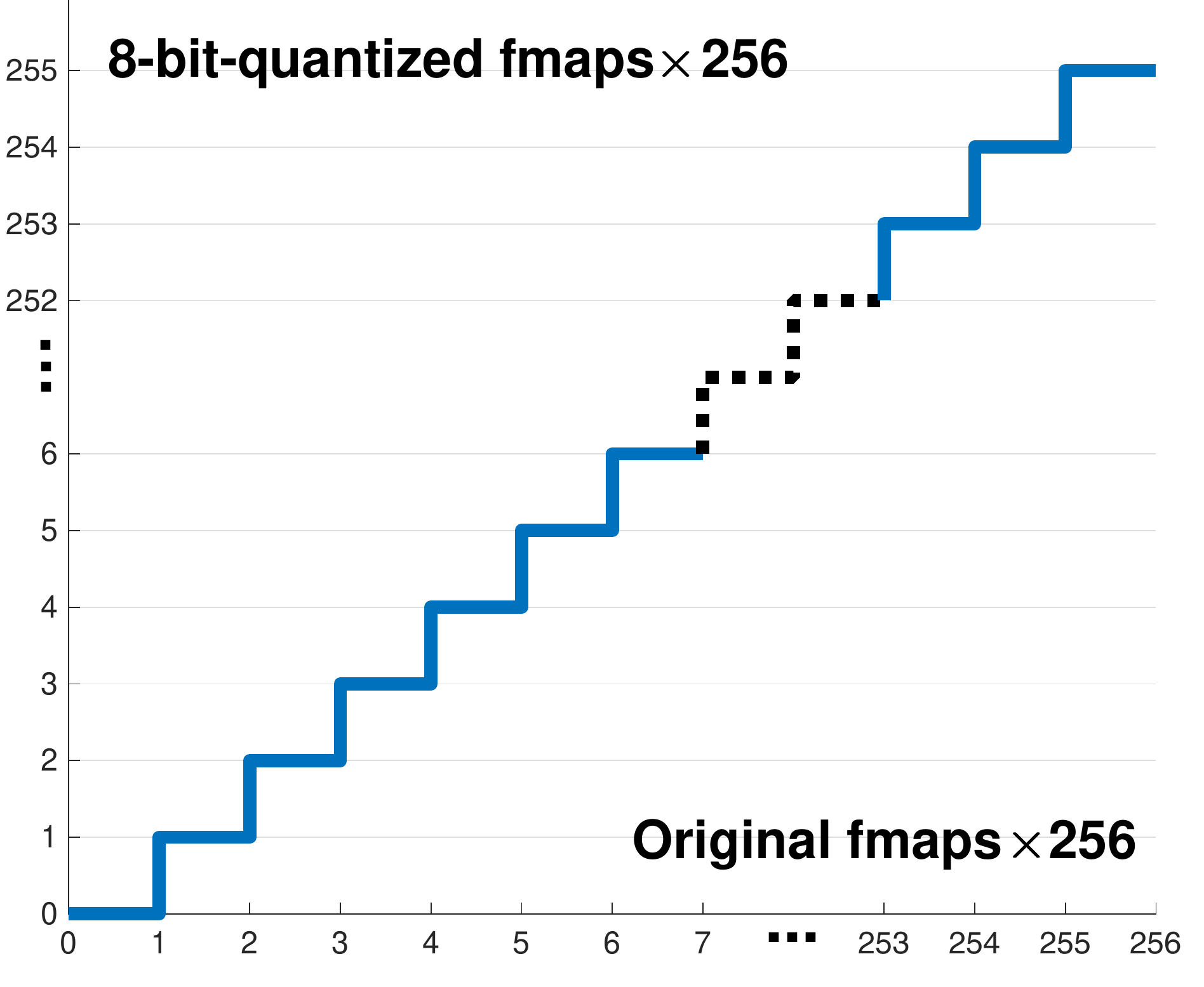}
		\label{fig:fmap}
	}
	\caption{8-bit quantization strategy of (a) weight and (b) feature map.}
	\label{fig:8bit}
\end{figure}

The convolution is the core operation of the YOLO and other CNN-based networks.
According to the recent works \cite{zhou2016dorefa, hubara2016quantized}, we present the low-bit-width convolution with 8-bit quantization values for weights in order to avoid the dropout of accuracy and also improve the performance.
Assuming that $w \in [-1 ,0]\cup [0, 1]$ are the full-precision weights and $w^q \in [-2^7+1, 0] \cup [0,2^7-1]$ are the 8-bit quantized-valued weights, and they have the approximation as $w \approx \xi \cdot w^q $ with a with a non-negative
scaling factor $\xi$. The weights are quantized in 8-bit as following:
\begin{equation}
w_i^q = quantize_w(w_i) =
\begin{cases}
\frac{w_i}{|w_i|},\ \ \ \ \ \ \ \ \ \ \ \ \ \ \ \ \ \ \ \ \ 0 < |w_i| \leq \frac{1}{2^7},\\
INT(2^{7}\times w_i), \ \ \ \frac{1}{2^{7}} < |w_i| < 1, \\
(2^{7}-1)\frac{w_i}{|w_i|},\ \ \ \ \ \ \ |w_i| = 1,\\
\end{cases}\\
\end{equation}
where the function $INT$ takes the smaller nearest integer.

Also we develop an activation with quantization which quantizes a real number feature maps $a \in [0, 1]$ to an 8-bit feature maps $a^q \in [0,1]$. This strategy is defined as below:
\begin{equation}
a_i^q = quantize_a(a_i) =
\frac{1}{2^{8}} \times
\begin{cases}
INT(2^{8}\times a_i), \ \ 0\leq a_i<1, \\
2^8-1,\ \ \ \ \ \ \ \ \ \ \ \ \ \ \ a_i = 1.\\
\end{cases}\\
\end{equation}

The detail distribution of 8-bit weights and feature maps is presented in Fig.\ref{fig:8bit}. Having both the quantized weights and feature maps, we can get the quantized convolution as assumed:
\begin{equation}
\label{eq:convolution}
		s_k^q(x,y,z) = \sum_{i=1}^{W_k}{\sum_{j=1}^{H_k}{\sum_{l=1}^{D_k}{w_k^q(i,j,l,z)}}} \cdot a_{k-1}^q(i+x-1, j+y-1, l),
\end{equation}
where $w_k^q$, $a_{k-1}^q\in {W_k\times H_k\times D_k}$ are the 8-bit-quantized weights and feature maps, respectively. Since the elements of weights and feature maps can be calculated and stored in 8-bit, both of the processor and memory resources required for quantized convolutional layer can be greatly reduced.

\begin{figure}[htb]
  \centering
  \includegraphics[width=0.8\textwidth]{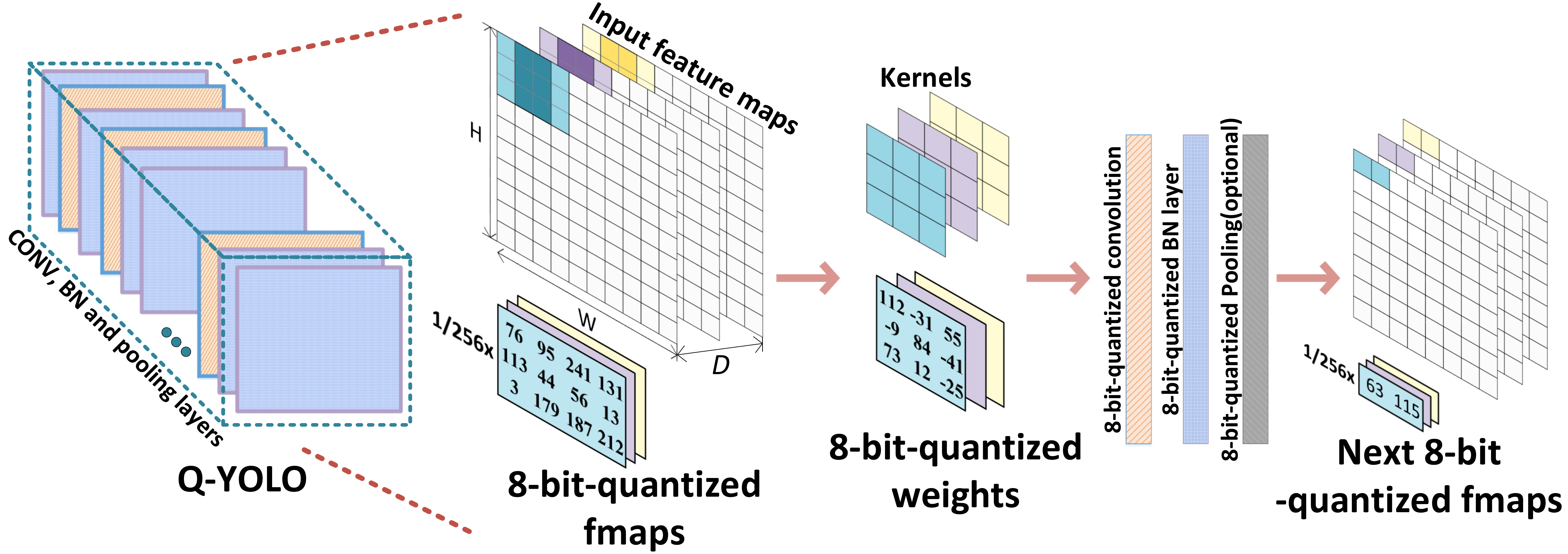}
  \caption{Q-YOLO: computation flow of forward propagation after 8-bit quantization.}
  \label{fig:qyolo}
\end{figure}

The overall working flow of the Q-YOLO model is presented in Fig.\ref{fig:qyolo}, and the network is assumed to have a feed-forward linear topology. We can have the observation that all the expensive operations in convolutional layers are operating on 8-bit quantization. The batch normalization layers and max-pooling layers are also quantized as 8-bit.

%% file: secs/ttrnn.tex
\section{RNN with Tensorization}
\label{sec:ttrnn}

Previous neural network compression on RNN is performed
by either precision-bit truncation or low-rank approximation
\cite{sainath2013low, denton2014exploiting, denil2013predicting}, which cannot maintain good balance between network compression and network accuracy. In this section, we discuss a tensorization-based
RNN during the training process. The tensor
decomposition method will be first introduced, and then
a tensorized RNN (namely T-RNN) will be discussed based on the extension of general neural network.

\subsection{Tensor Decomposition}
\label{sec:tt}
Tensors are natural multi-dimensional generation of matrices, and the tensor-train factorization \cite{ye2017learning, tjandra2018tensor} is a promising tensorial decomposition model that can scale to an arbitrary number of dimensions.
We refer one-dimensional data as vectors, denoted
as $a$, two-dimensional arrays are matrices, denoted as $A$ and
the higher dimensional arrays are tensors denoted as $\mathcal{A}\in \mathbb{R}^{l_1\times l_2 \times \ldots \times l_d}$, $\mathcal{A}(h) = \mathcal{A}(h_1, h_2, \ldots, h_d)$ (refer one specific element from a tensor using calligraphic upper
letters), where $d$ is the dimensionality
of the tensor.

The d-dimensional tensor $\mathcal{A}$ can be decomposed
by using the tensor core $\mathcal{G}_k \in \mathbb{R}^{l_k\times r_{k}\times r_{k-1}}$ and each element $\mathcal{A}(h_1, h_2, \ldots, h_d)$  is defined as:
\begin{equation}
\label{eq:tensor1}
\mathcal{A}(h_1, h_2, \ldots, h_d)  = \sum_{\alpha_0,\alpha_1,\ldots, \alpha_d}^{r_0,r_1,\ldots, r_d} \mathcal{G}_1(h_1,\alpha_0,\alpha_1) \mathcal{G}_2(h_2,\alpha_1,\alpha_2)\ldots \mathcal{G}_d(h_d,\alpha_{d-1},\alpha_d)
\end{equation}
where $\alpha_k$ is the index of summation which starts from $1$ and
stops at rank $r_k$. It should be noted that $r_0 = r_d = 1$ for the boundary condition
and $l_1, l_2, \ldots, l_d$ are known as mode size. Here, $r_k$ is the core
rank and $\mathcal{G}$ is the core for this tensor decomposition. By using
the notation of $\mathcal{G}_k(h_k) \in \mathbb{R}^{r_k \times r_{k-1}}$ (a 2-dimensional slice from the 3-dimensional tensor $\mathcal{G}_k$), we can rewrite the above
equation in a more compact way:
\begin{equation}
\label{eq:tensor2}
\mathcal{A}(h_1, h_2, \ldots, h_d)  = \mathcal{G}_1(h_1) \mathcal{G}_2(h_2)\ldots \mathcal{G}_d(h_d)
\end{equation}

Imposing the constraint that each integer $l_k$ as shown in Eq.\ref{eq:tensor2} can be factorized as $l_k = m_k \cdot n_k, \forall k\in [1,d]$, and consequently
reshapes each $\mathcal{G}_k$ into $\mathcal{G}_k^* \in \mathbb{R}^{m_k \times n_k \times r_k \times r_{k-1}}$.
The decomposition for the tensor $\mathcal{A}\in \mathbb{R}^{(m_1\cdot n_1)\times (m_2\cdot n_2) \times \ldots \times (m_d\cdot n_d)}$ can be correspondingly reformulated as:
\begin{equation}
\label{eq:tensor3}
\mathcal{A}((i_1, j_1), (i_2, j_2), \ldots, (i_d, j_d))  = \mathcal{G}^*_1(i_1, j_1) \mathcal{G}^*_2(i_2, j_2)\ldots \mathcal{G}^*_d(i_d, j_d)
\end{equation}

This double index trick \cite{novikov2015tensorizing} enables the
factorizing of the computing in a fully-connected layer, which will be discussed in following section.

\subsection{Tensorized RNN}

The core operation in RNN is fully-connected layer and its computing process can be compactly described as:
\begin{equation}
y(j) = \sum_{i=1}^M W(i,j) \cdot x(i) + b(j)
\end{equation}
where $x \in \mathbb{R}^M$, $y \in \mathbb{R}^N$ and $j \in [1,N]$. Assuming that $M = \Pi_{k=1}^d m_k$, $N = \Pi_{k=1}^d n_k$, we can reshape the tensors $x$ and $y$ into tensors with d-dimension: $\mathcal{X} \in \mathbb{R}^{m_1 \times m_2 \times \ldots \times m_d}$, $\mathcal{Y} \in \mathbb{R}^{n_1 \times n_2 \times \ldots \times n_d}$, and then the fully-connected computing function is turning out to be:
\begin{equation}
\label{eq:tensor5}
\begin{split}
\mathcal{Y}(j_1, j_2, \ldots, j_d) =& \sum_{i_1=1}^{m_1}\sum_{i_2=1}^{m_2}\ldots\sum_{i_d=1}^{m_d} [\mathcal{W}((i_1,j_1),(i_2,j_2),\ldots,(i_d,j_d)) \\
&\times \mathcal{X}(i_1,i_2,\ldots,i_d)] + \mathcal{B}(j_1,j_2,\ldots,j_d)
\end{split}
\end{equation}

\begin{figure}[htb]
  \centering
  \includegraphics[width=0.95\textwidth]{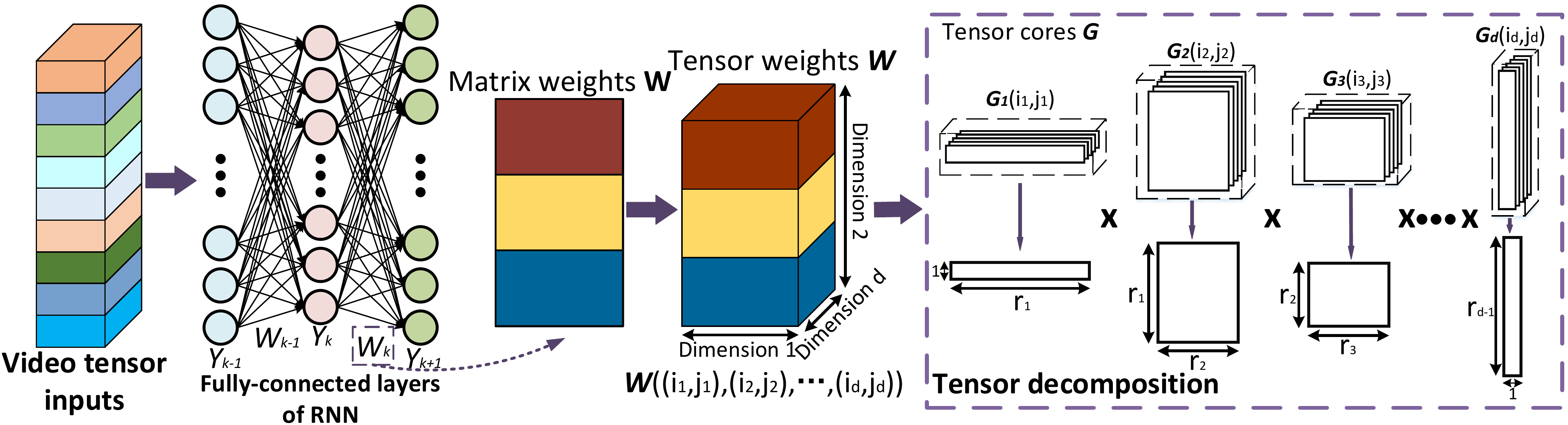}
  \caption{T-RNN: tensorized weights for parameter compression.}
  \label{fig:tt}
\end{figure}

The whole working flow with tensorization on the hidden-to-hidden weights is shown in Fig.\ref{fig:tt}. Due to the above decomposition in Eq.\ref{eq:tensor3}, the calculating multiplication complexity turns out to be $O(dr^2n_m)$ \cite{novikov2015tensorizing} instead of $O(n^d)$, where $r$ is the maximum
rank of cores $\mathcal{G}_k$ and $n_m$ is the maximum mode size $m_k \cdot n_k$ of
tensor $\mathcal{W}$. This will be much higher compressed and more efficient since the rank $r$ is very small
compared with general matrix-vector multiplication of traditional fully-connected layers.

%
%

%% file: secs/overall.tex
\section{DEEPEYE Framework for Video Comprehension}
\label{sec:deepeye}

\begin{figure}[htb]
  \centering
  \includegraphics[width=0.9\textwidth]{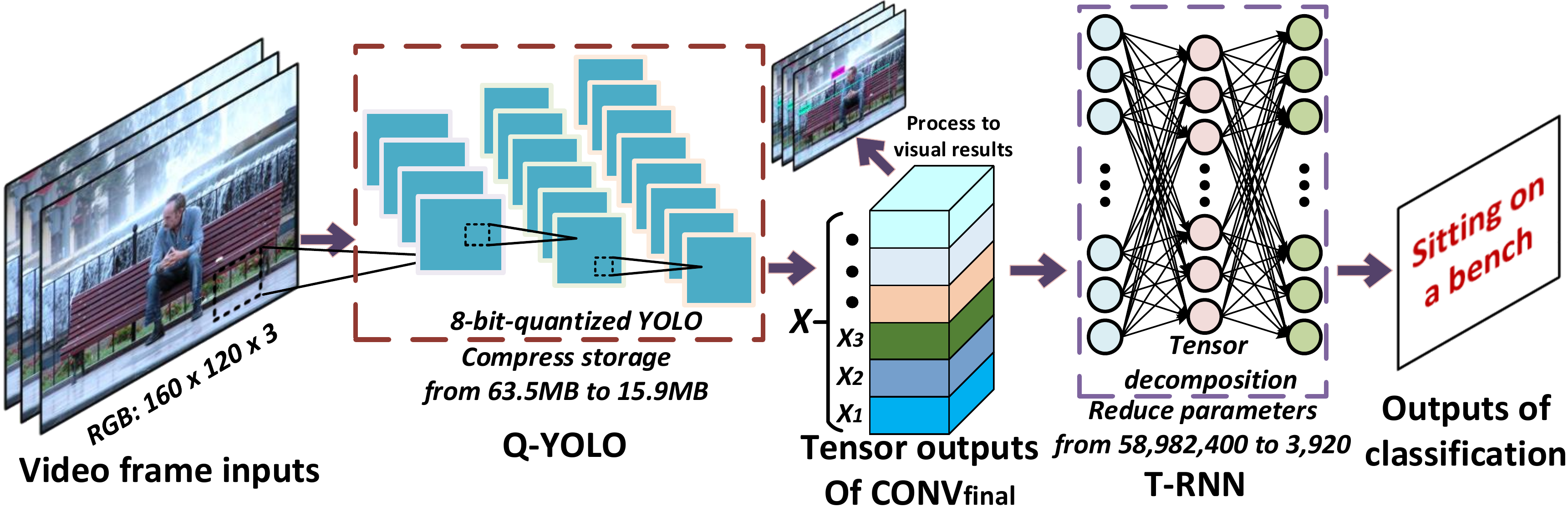}
  \caption{DEEPEYE: detailed real-time working flow composed by Q-YOLO and T-RNN.}
  \label{fig:overall}
\end{figure}

Based on the quantization and tensorization of YOLO, the whole working flow of DEEPEYE framework for video comprehension is shown in Fig.\ref{fig:overall}. It integrates the Q-YOLO, served as real-time video object detection and the T-RNN, served as video classification system. Firstly, the prepared video clip is primarily delivered into Q-YOLO as inputs, where all the convolutional layers, batch normalization layers and max-pooling layers are quantized as 8-bit. Then, the tensor feature outputs of Q-YOLO can be further fed to T-RNN without delay. It should be noted that the tensor feature outputs are the final results of the last convolution layer ($CONV_{final}$) from Q-YOLO which can also be further processed to display the real-time visual results. Finally, after the T-RNN processing with tensorized-compression on both tensorial input-to-hidden and hidden-to-hidden mappings, one can obtain the classification result towards action recognition.

Below, we further summarize the training steps of DEEPEYE as follows:


\begin{enumerate}
\item \textbf{Train Q-YOLO:} Train the Q-YOLO with existing or customized dataset for object detection.
    The feature outputs of $CONV_{final}$ in Q-YOLO for each frame are tensor format data $\mathcal{X}(i_1, i_2,\ldots, i_d) \in \mathbb{R}^{m_1 \times m_2 \times \ldots \times m_d}$, and can be represented by subitems $\mathcal{X}(i_1) \times \mathcal{X}(i_2)\times \ldots \times \mathcal{X}(i_d)$ such as $19 \times 19 \times 425$ in the experiments.

\item \textbf{Pre-process video dataset:} Pre-process the existing or customized video dataset (in the experiments MOMENTS \cite{monfort2018moments} and UCF11 \cite{liu2009recognizing} are used) with Q-YOLO. Fed each video clip to Q-YOLO to obtain its tensor outputs $\mathcal{X}$, which are regarded as the tensor format dataset for T-RNN instead of the original frame format dataset.


\item \textbf{Train T-RNN:} Train the model with the tensor format dataset and tensorized weights $\mathcal{W}$.
    The final output model will be used for a real-time classification.

\item \textbf{Understand in real time:} After both the Q-YOLO and T-RNN models have been trained, the whole video comprehension flow can be built for real time analysis as shown in Fig.\ref{fig:overall}.
\end{enumerate}

Instead of optimizing the video detection and classification in a separated fashion, the DEEPEYE is the first approach to leverage object detection and action recognition together with remarkable optimizations.
Since the whole system is highly compressed with quantization and tensorization, it benefits a lot with a much better performance in compression, speed-up as well as resource-saving, especially when applying to the video comprehension tasks.
As presented in Fig.\ref{fig:overall}, the storage cost of the experiments is compressed from $63.5 MB$ to $15.9 MB$ and the number of parameters is reduced from $58,982,400$ to $3920$, and detail layers model of proposed Q-YOLO is presented in Table \ref{tbl:layer}.
\begin{table}[htb]
\centering
\caption{The framework of layers model in DEEPEYE: the memory size is listed with the storage costs before and after 8-bit quantization separately.}
\label{tbl:layer}
\scalebox{1}{
\begin{tabular}{ccccc}
\hline
\hline
Layer      & Filters  & Output & Parameters & Memory Size   \\ \hline
$CONV_1$     & $608\times608 \times3$     & $608\times608 \times16$  & $432$ & $1.7$KB, $0.4$KB   \\ \hline
$CONV_2$      & $608\times608 \times16$     & $304\times304 \times32$  & $4,608$ & $17.9$KB, $4.5$KB \\ \hline
$CONV_3$       & $152\times152 \times32$     & $152\times152 \times64$  & $18,432$ & $71.9$KB, $18.1$KB \\ \hline
...     & ...  & ... & ... & ... \\ \hline
$CONV_{final}$   & $19\times19 \times1024$     & $19\times19 \times125$  & $128,000$ & $499.3$KB, $125.0$KB \\ \hline \hline
\end{tabular}
}
\end{table}


%% file: secs/result.tex
\section{Experiments}
\label{sec:result}

In the experiments, we have implemented different baselines for performance comparison as follows. $\textbf{1,}$ DEEPEYE: Proposed video comprehension system combining the tiny-YOLOv2 with quantization (Q-YOLO) and LSTM (the advanced variant of RNN) with tensorization (T-RNN). We apply $0.25$ dropout \cite{srivastava2014dropout} for both input-to-hidden and hidden-to-hidden mappings in T-RNN. $\textbf{2,}$ Original YOLO: The original full-precision tiny-YOLOv2 without quantization and only for video detection. $\textbf{3,}$ Plain RNN: The plain RNN without tensorization and only for video classification, which inputs are the original video frame data instead of the tensor outputs of $CONV_{final}$ in Q-YOLO. $\textbf{4,}$ T-RNN with frame inputs: T-RNN with original inputs of video frame is also selected for performance comparison.

%

It should be noted that all the baselines are implemented in the same initialization environment: Theano 
in Keras for software and NVIDIA GTX-1080Ti for hardware.
We validate the contributions of our system by presenting a comparison study
on two challenging large video datasets (MOMENTS \cite{monfort2018moments} and UCF11 \cite{liu2009recognizing}), as discussed in the following sections.

\subsection{Comparison on Video Detection}
\label{sec:detection}

To show the effects of video detection, we apply the MOMENTS dataset which contains one million labeled $3$ second video clips, involving people, animals, objects or natural phenomena, that capture the gist of a dynamic scene.
Each clip is assigned with $339$ action classes such as eating, bathing or attacking.
Based on
the majority of the clips we resize every frames to a standard size $340 \times 256$, at the
fps $25$. For a premier experiment, we choose representational $10$ classes and the length of training sequences is set to be $80$ while the length of test
sequences is $20$.

\begin{figure}[htb]
	\centering
	\subfigure[]
	{
		\includegraphics[width=0.31\textwidth]{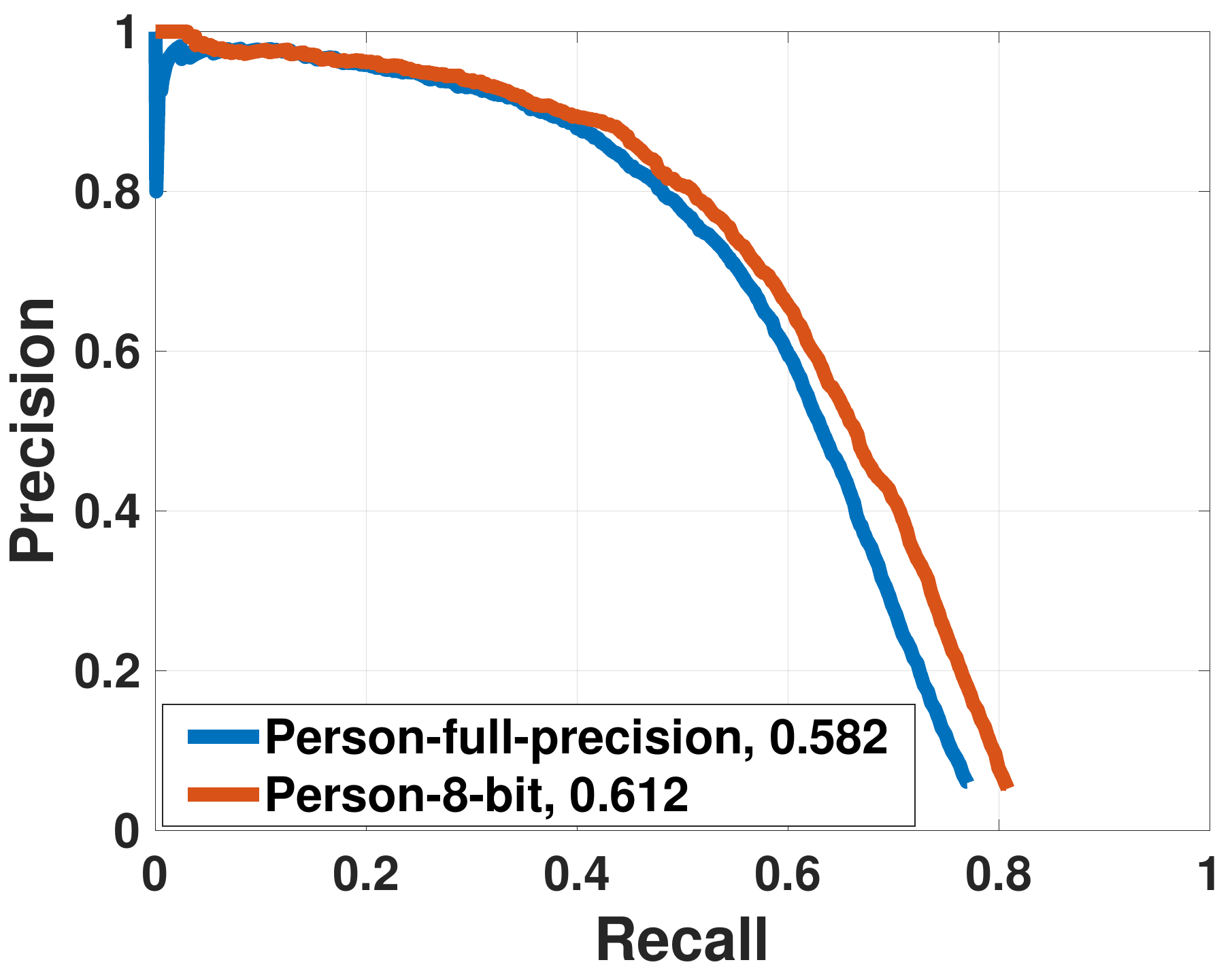}
		\label{fig:person}
	}
	\subfigure[]
	{
		\includegraphics[width=0.31\textwidth]{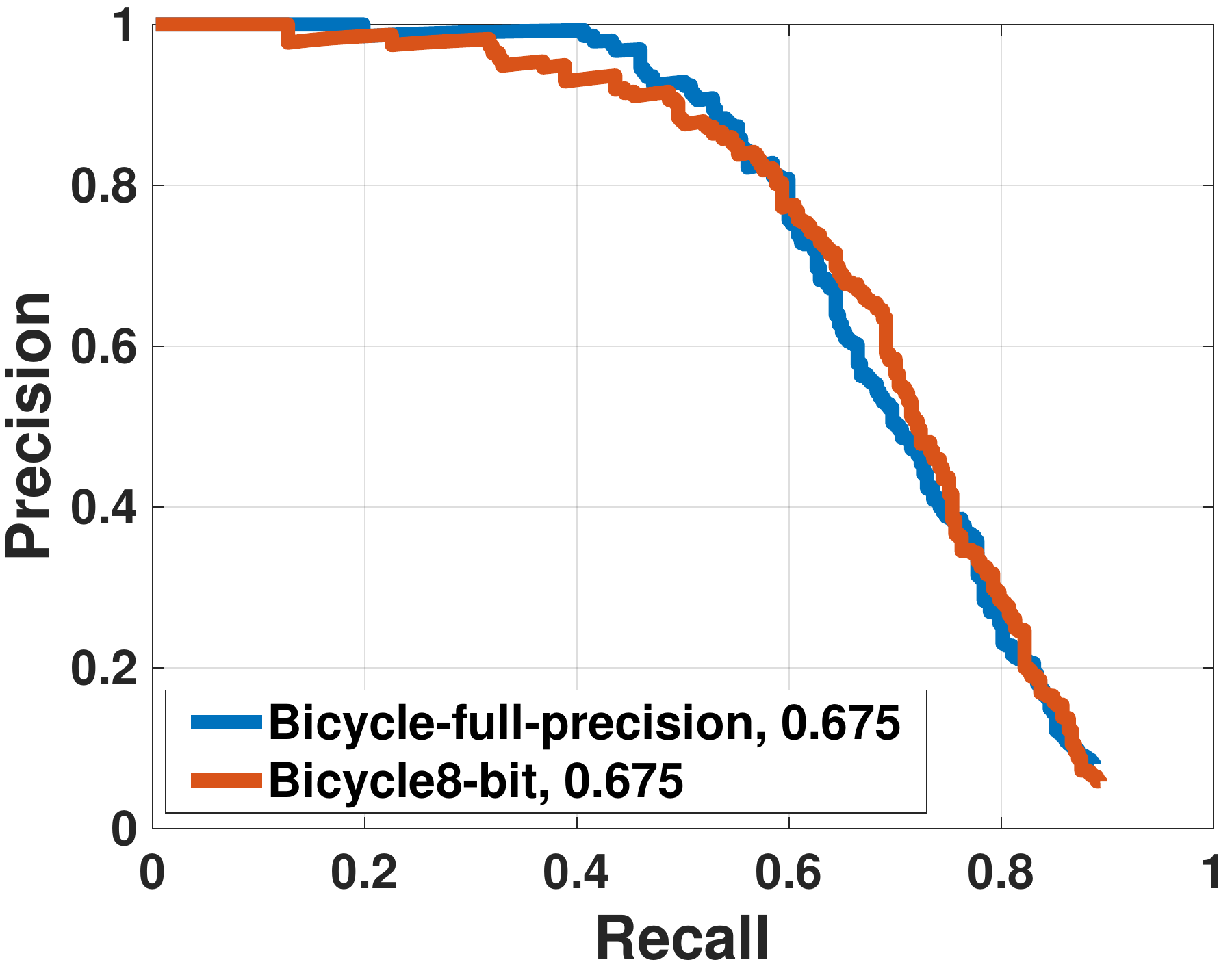}
		\label{fig:bicycle}
	}
	\subfigure[]
	{
		\includegraphics[width=0.31\textwidth]{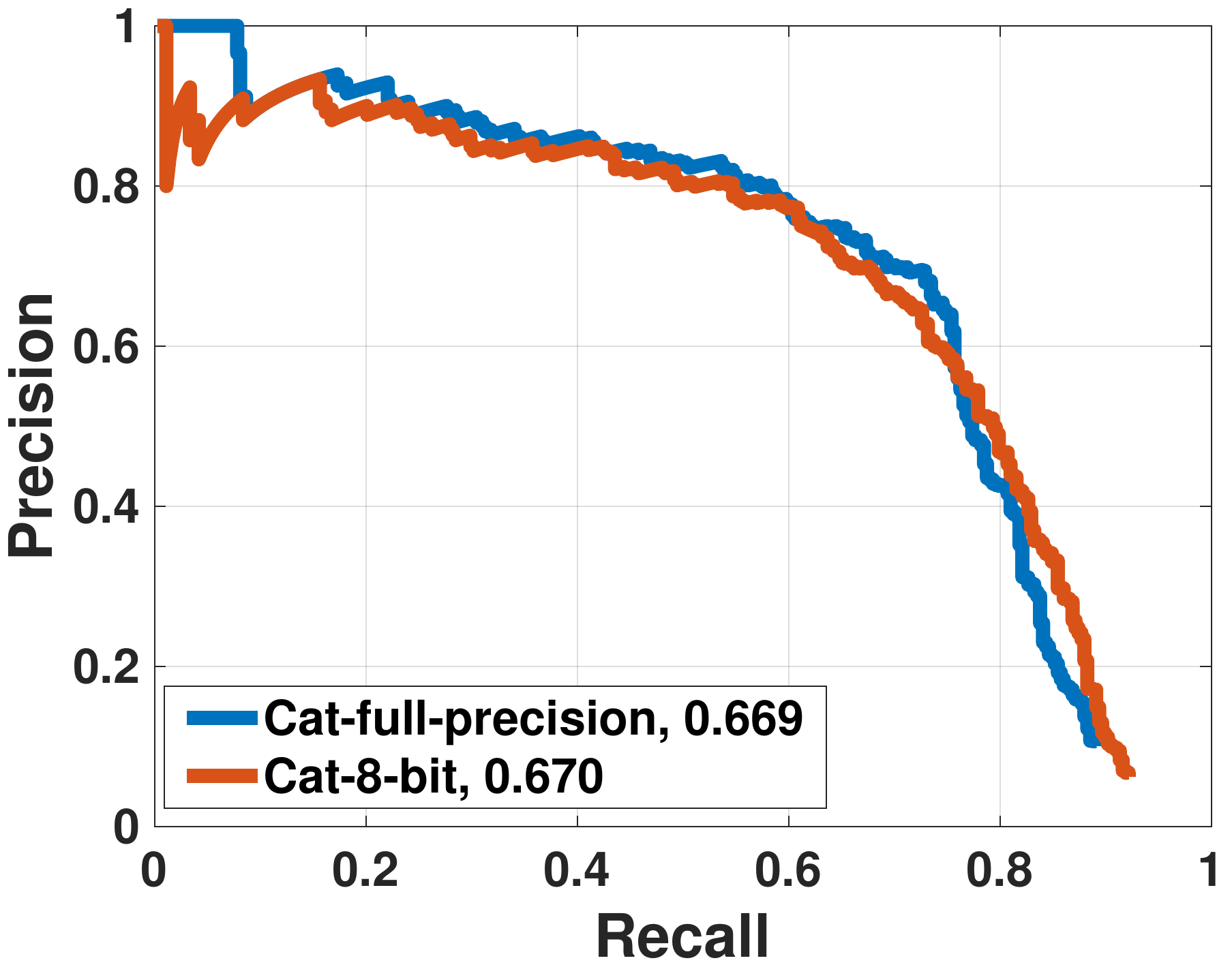}
		\label{fig:cat}
	}
	\caption{Average precision comparison between 8-bit-quantized Q-YOLO and full-precision YOLO
on classes: (a) person, (b) bicycle and (c) cat.}
	\label{fig:ap}
\end{figure}

Firstly, we pre-train the Q-YOLO on VOC with $20$ object classes. As shown in Fig.\ref{fig:ap}, we report the Average Precision (AP) comparison between the proposed 8-bit-quantized model and full-precision model on $3$ representational classes (the AP score corresponds to the Area-Under-Precision-Recall-Curve).
Then, the mean Average Precision (mAP) among all $20$ classes is obtained, which can reach $0.5350$ in the 8-bit Q-YOLO while the mAP of full-precision YOLO is $0.5397$.
It can be seen that the 8-bit Q-YOLO does not cause the AP curves to be significantly different from full-precision one and only $0.47\%$ decreases on mAP. As such, we can see that the Q-YOLO with 8-bit quantization obtains a commendable balance between large compression and high accuracy.

\begin{figure}[htb]
  \centering
  \includegraphics[width=0.98\textwidth]{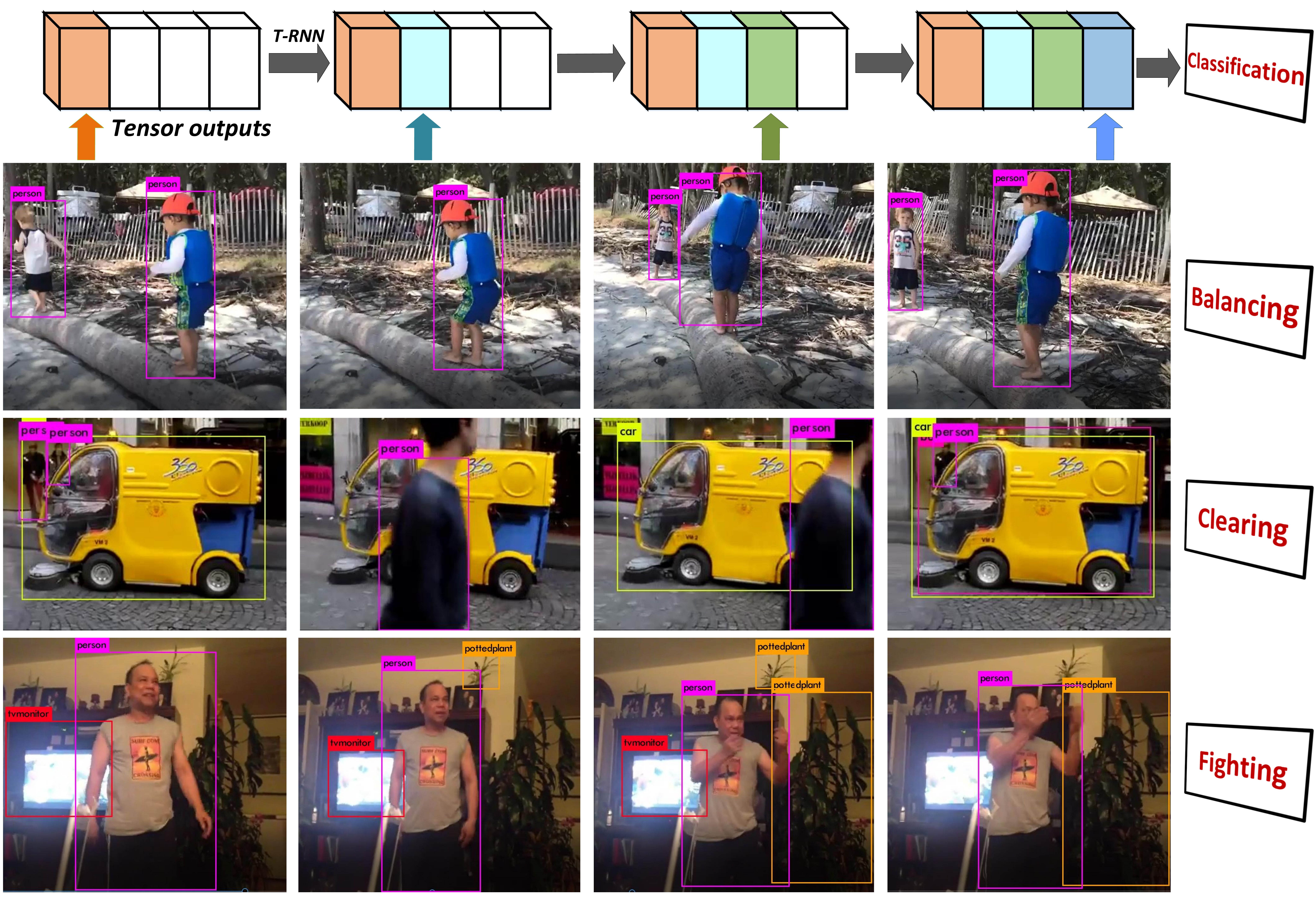}
  \caption{Visual results of DEEPEYE on MOMENTS dataset.}
  \label{fig:detection}
\end{figure}

Secondly, the visual results of our approach on MOMENTS are shown in Fig.\ref{fig:detection}. Experimental results show that all existing objects in these video clips can be detected precisely in real time. In this system, the finally tensor output of each frame is in a size of $19 \times 19 \times 425$, which is delivered into T-RNN for video classification with no delay.

\subsection{Comparison on Video Classification}

In this section, we use UCF11 dataset for a performance comparison on video classification. The dataset contains $1600$ video clips, falling
into $11$ action classes that summarize
the human action visible in each video clip such as
basketball shooting, biking or diving. We resize the RGB frames into $160 \times 120$ at the fps $24$.

\begin{figure}[htb]
	\centering
	\subfigure[]
	{
		\includegraphics[width=0.48\textwidth]{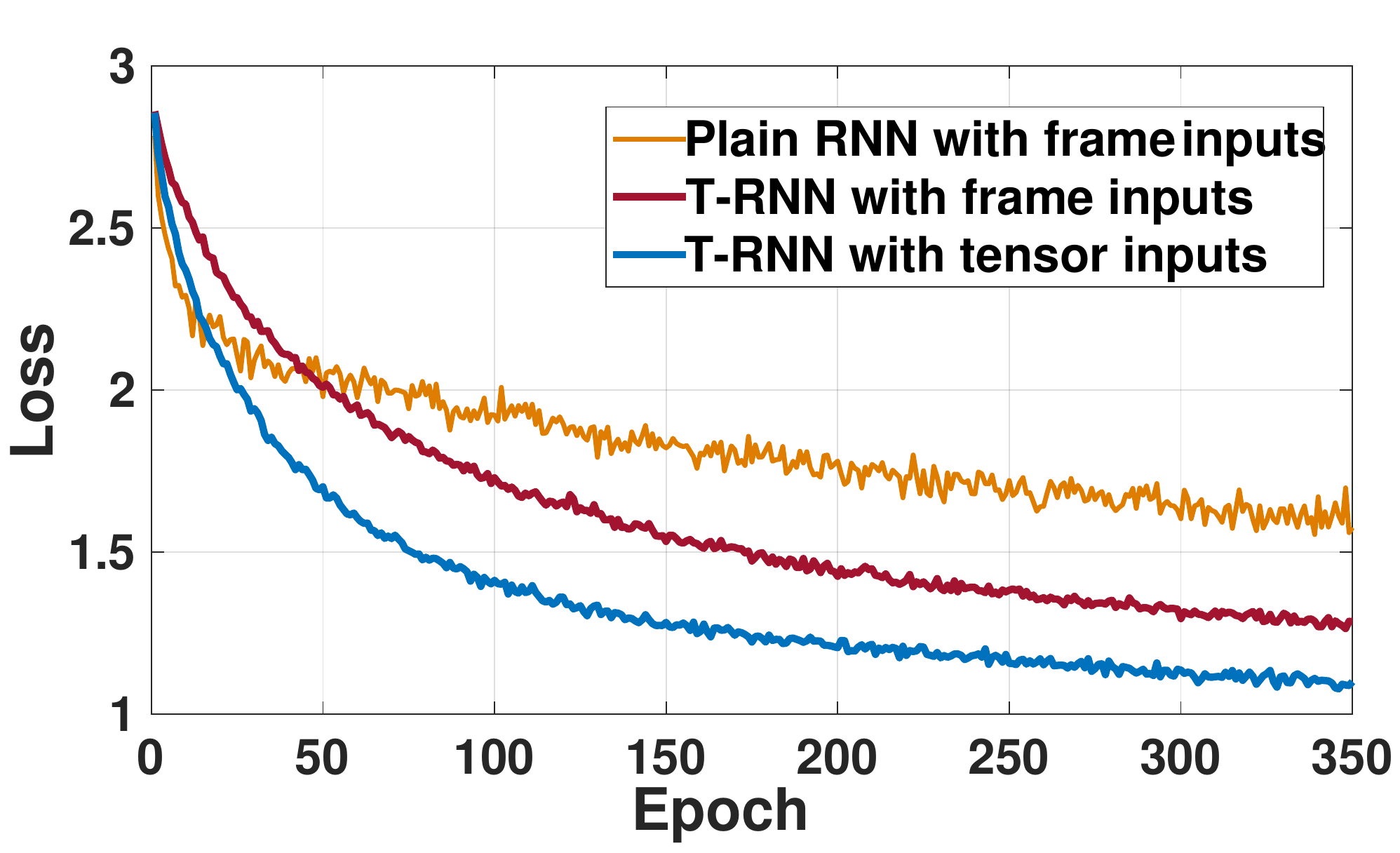}
		\label{fig:loss}
	}
	\subfigure[]
	{
		\includegraphics[width=0.48\textwidth]{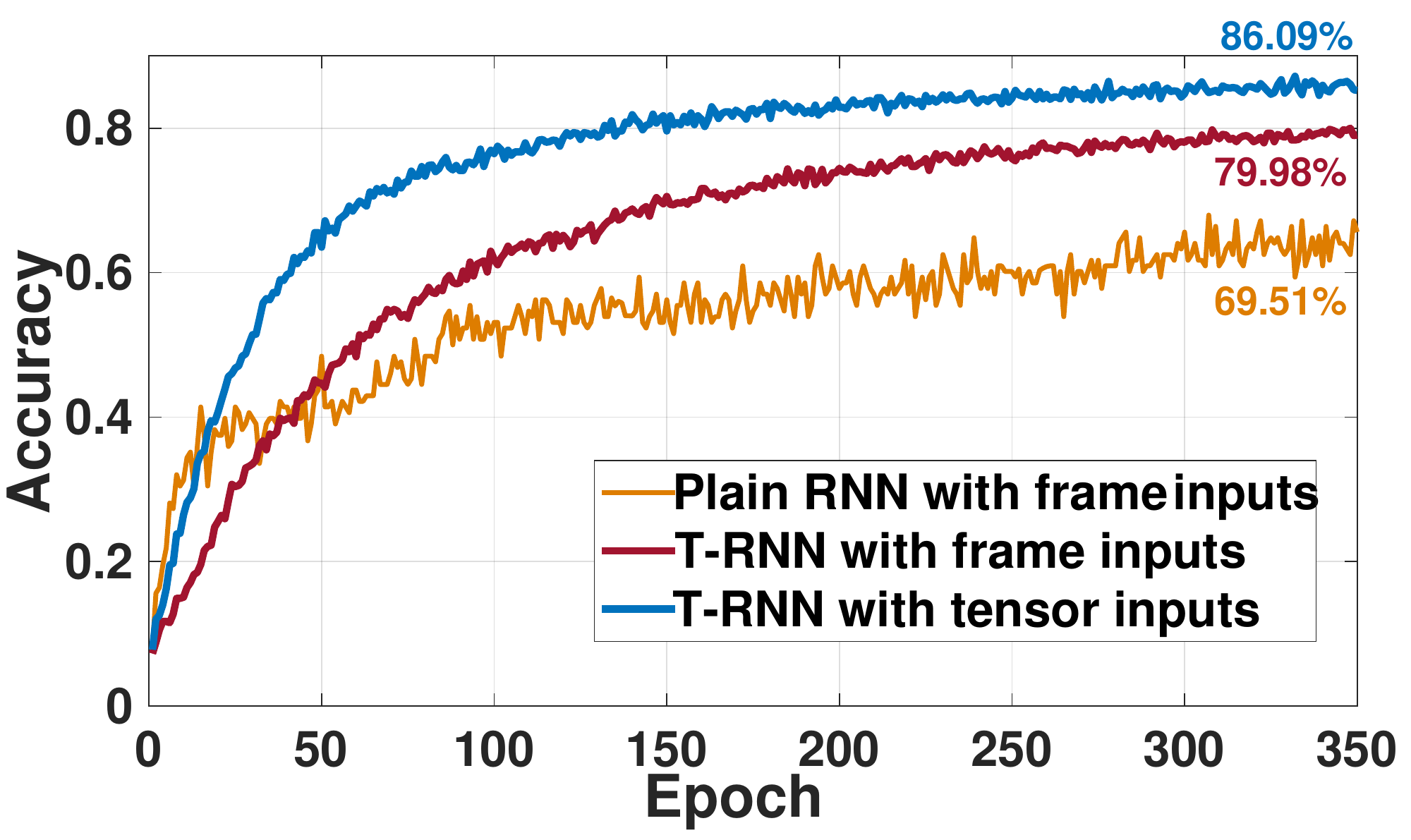}
		\label{fig:acc}
	}
	\caption{Comparison among different RNN models on (a) loss curves and (b) accuracy curves.}
	\label{fig:classification}
\end{figure}

We sample $6$ random frames in ascending order from each video clip as the input data \cite{yang2017tensor}. The tensorization-based algorithm has been configured for both inputs and weights by the training process. Fig.\ref{fig:classification} shows the training loss and accuracy comparison among: $\textbf{1)}$ T-RNN with tensor inputs (Q-YOLO outputs), $\textbf{2)}$ T-RNN with frame inputs and $\textbf{3)}$ plain RNN with frame inputs. We set the parameters as follows: the tensor dimension is $d = 4$; the shapes
of inputs tensor are apart and we set them as: $\textbf{1)}$ $ m_1 = 17, m_2 = 19, m_3 = 19, m_4 = 25$, $\textbf{2)}$ $ m_1 = 8, m_2 = 20, m_3 = 20, m_4 = 18$;
the hidden shapes are $n_1 = n_2 = n_3 = n_4 = 4$; and the ranks of
T-RNN are $r_1 = r_5 = 1$, $r_2 = r_3 = r_4 = 4$.

It can be seen that when the T-RNN with tensor format inputs performs the best once the epoch beyond $20$. The peak accuracy of proposed framework reaches $86.09\%$, $16.58\%$ higher than the plain RNN while $6.11\%$ higher than the T-RNN with frame inputs, which tremendously improve the accuracy performance.

\subsection{Performance Analysis}

Aside from the outstanding function and accuracy, the high compression and speed-up are also remarkable.
The proposed DEEPEYE can cost $3.994 \times$ less storage and computing resources compared with the full-precision YOLO. Since the complexity is significantly reduced and the throughput of networks is highly enlarged, the future implementation on terminal devices also becomes more realizable. The performance evaluation is shown in Table. \ref{tbl:result} based on different baselines. Among all the baselines, the proposed DEEPEYE system (T-RNN with tensor inputs) has the most excellent performance which deliveries much better accuracies even with several orders of less parameters.

\begin{table}[htb]
\centering
\caption{
The performance evaluation on multiple baselines: the number of parameters involved in
the input-to-hidden mappings of respective baselines; and the runtime as average cost-time of
each training epoch.
}
\label{tbl:result}
\scalebox{1}{
\begin{tabular}{cccccc}
\hline
\hline
      & Accuracy & Parameters & Compression & Runtime & Speed-up  \\ \hline
RNN        & $69.51\%$      & $58,982,400$    & -    & $1,369 s$ & -     \\ \hline
T-RNN with frame inputs & $79.98\%$  & $3,360$ & $17,554 \times$ & $431 s$ & $3.18 \times$   \\ \hline
DEEPEYE & $86.09\%$ & $3,920$ & $15,047 \times$ & $477 s$ & $2.87 \times$    \\ \hline \hline
\end{tabular}
}
\end{table}

%% file: nips_2018.bbl
\begin{thebibliography}{10}
\providecommand{\url}[1]{#1}
\csname url@samestyle\endcsname
\providecommand{\newblock}{\relax}
\providecommand{\bibinfo}[2]{#2}
\providecommand{\BIBentrySTDinterwordspacing}{\spaceskip=0pt\relax}
\providecommand{\BIBentryALTinterwordstretchfactor}{4}
\providecommand{\BIBentryALTinterwordspacing}{\spaceskip=\fontdimen2\font plus
\BIBentryALTinterwordstretchfactor\fontdimen3\font minus
  \fontdimen4\font\relax}
\providecommand{\BIBforeignlanguage}[2]{{%
\expandafter\ifx\csname l@#1\endcsname\relax
\typeout{** WARNING: IEEEtran.bst: No hyphenation pattern has been}%
\typeout{** loaded for the language `#1'. Using the pattern for}%
\typeout{** the default language instead.}%
\else
\language=\csname l@#1\endcsname
\fi
#2}}
\providecommand{\BIBdecl}{\relax}
\BIBdecl

\bibitem{lecun1998gradient}
Y.~LeCun, L.~Bottou, Y.~Bengio, and P.~Haffner, ``Gradient-based learning
  applied to document recognition,'' \emph{Proceedings of the IEEE}, vol.~86,
  no.~11, pp. 2278--2324, 1998.

\bibitem{krizhevsky2012imagenet}
A.~Krizhevsky, I.~Sutskever, and G.~E. Hinton, ``Imagenet classification with
  deep convolutional neural networks,'' in \emph{Advances in Neural Information
  Processing Systems}, 2012, pp. 1097--1105.

\bibitem{redmon2018yolov3}
J.~Redmon and A.~Farhadi, ``Yolov3: An incremental improvement,'' \emph{arXiv
  preprint arXiv:1804.02767}, 2018.

\bibitem{guo2017software}
K.~Guo, S.~Han, S.~Yao, Y.~Wang, Y.~Xie, and H.~Yang, ``Software-hardware
  codesign for efficient neural network acceleration,'' \emph{IEEE Micro},
  vol.~37, no.~2, pp. 18--25, 2017.

\bibitem{hashemi2017understanding}
S.~Hashemi, N.~Anthony, H.~Tann, R.~I. Bahar, and S.~Reda, ``Understanding the
  impact of precision quantization on the accuracy and energy of neural
  networks,'' in \emph{Design, Automation \& Test in Europe Conference \&
  Exhibition (DATE)}, 2017, pp. 1474--1479.

\bibitem{courbariaux2015binaryconnect}
M.~Courbariaux, Y.~Bengio, and J.-P. David, ``Binaryconnect: Training deep
  neural networks with binary weights during propagations,'' in \emph{Advances
  in Neural Information Processing Systems}, 2015, pp. 3123--3131.

\bibitem{liu2018squeezedtext}
Z.~Liu, Y.~Li, F.~Ren, H.~Yu, and W.~Goh, ``Squeezedtext: A real-time scene
  text recognition by binary convolutional encoder-decoder network,'' 2018.

\bibitem{redmon2016you}
J.~Redmon, S.~Divvala, R.~Girshick, and A.~Farhadi, ``You only look once:
  Unified, real-time object detection,'' in \emph{Proceedings of the IEEE
  Conference on Computer Vision and Pattern Recognition}, 2016, pp. 779--788.

\bibitem{redmon2017yolo9000}
J.~Redmon and A.~Farhadi, ``Yolo9000: better, faster, stronger,'' \emph{arXiv
  preprint arXiv:1612.08242}, 2017.

\bibitem{yao2015describing}
L.~Yao, A.~Torabi, K.~Cho, N.~Ballas, C.~Pal, H.~Larochelle, and A.~Courville,
  ``Describing videos by exploiting temporal structure,'' in \emph{Proceedings
  of the IEEE Conference on Computer Vision}, 2015, pp. 4507--4515.

\bibitem{ebrahimi2015recurrent}
S.~Ebrahimi~Kahou, V.~Michalski, K.~Konda, R.~Memisevic, and C.~Pal,
  ``Recurrent neural networks for emotion recognition in video,'' in
  \emph{Proceedings of the ACM Conference on Multimodal Interaction}, 2015, pp.
  467--474.

\bibitem{venugopalan2015sequence}
S.~Venugopalan, M.~Rohrbach, J.~Donahue, R.~Mooney, T.~Darrell, and K.~Saenko,
  ``Sequence to sequence-video to text,'' in \emph{Proceedings of the IEEE
  Conference on Computer Vision}, 2015, pp. 4534--4542.

\bibitem{ng2015beyond}
J.~Y.-H. Ng, M.~Hausknecht, S.~Vijayanarasimhan, O.~Vinyals, R.~Monga, and
  G.~Toderici, ``Beyond short snippets: Deep networks for video
  classification,'' in \emph{Proceedings of the IEEE Conference on Computer
  Vision and Pattern Recognition}, 2015, pp. 4694--4702.

\bibitem{fernando2016learning}
B.~Fernando and S.~Gould, ``Learning end-to-end video classification with
  rank-pooling,'' in \emph{International Conference on Machine Learning}, 2016,
  pp. 1187--1196.

\bibitem{sharma2015action}
W.~Zhu, J.~Hu, G.~Sun, X.~Cao, and Y.~Qiao, ``A key volume mining deep
  framework for action recognition,'' in \emph{Proceedings of the IEEE
  Conference on Computer Vision and Pattern Recognition}, 2016, pp. 1991--1999.

\bibitem{srivastava2015unsupervised}
N.~Srivastava, E.~Mansimov, and R.~Salakhudinov, ``Unsupervised learning of
  video representations using lstms,'' in \emph{International Conference on
  Machine Learning}, 2015, pp. 843--852.

\bibitem{donahue2015long}
J.~Donahue, L.~Anne~Hendricks, S.~Guadarrama, M.~Rohrbach, S.~Venugopalan,
  K.~Saenko, and T.~Darrell, ``Long-term recurrent convolutional networks for
  visual recognition and description,'' in \emph{Proceedings of the IEEE
  Conference on Computer Vision and Pattern Recognition}, 2015, pp. 2625--2634.

\bibitem{yang2017tensor}
Y.~Yang, D.~Krompass, and V.~Tresp, ``Tensor-train recurrent neural networks
  for video classification,'' \emph{arXiv preprint arXiv:1707.01786}, 2017.

\bibitem{tjandra2018tensor}
S.~Zhe, K.~Zhang, P.~Wang, K.-c. Lee, Z.~Xu, Y.~Qi, and Z.~Ghahramani,
  ``Distributed flexible nonlinear tensor factorization,'' in \emph{Advances in
  Neural Information Processing Systems}, 2016, pp. 928--936.

\bibitem{everingham2010pascal}
M.~Everingham, L.~Van~Gool, C.~K. Williams, J.~Winn, and A.~Zisserman, ``The
  pascal visual object classes (voc) challenge,'' \emph{International Journal
  of Computer Vision}, vol.~88, no.~2, pp. 303--338, 2010.

\bibitem{zhou2016dorefa}
S.~Zhou, Y.~Wu, Z.~Ni, X.~Zhou, H.~Wen, and Y.~Zou, ``Dorefa-net: Training low
  bitwidth convolutional neural networks with low bitwidth gradients,''
  \emph{arXiv preprint arXiv:1606.06160}, 2016.

\bibitem{zhu2016trained}
C.~Zhu, S.~Han, H.~Mao, and W.~J. Dally, ``Trained ternary quantization,''
  \emph{arXiv preprint arXiv:1612.01064}, 2016.

\bibitem{hubara2016quantized}
I.~Hubara, M.~Courbariaux, D.~Soudry, R.~El-Yaniv, and Y.~Bengio, ``Quantized
  neural networks: Training neural networks with low precision weights and
  activations,'' \emph{arXiv preprint arXiv:1609.07061}, 2016.

\bibitem{sainath2013low}
T.~N. Sainath, B.~Kingsbury, V.~Sindhwani, E.~Arisoy, and B.~Ramabhadran,
  ``Low-rank matrix factorization for deep neural network training with
  high-dimensional output targets,'' in \emph{Proceedings of the IEEE
  Conference on Acoustics, Speech and Signal Processing}, 2013, pp. 6655--6659.

\bibitem{denton2014exploiting}
E.~L. Denton, W.~Zaremba, J.~Bruna, Y.~LeCun, and R.~Fergus, ``Exploiting
  linear structure within convolutional networks for efficient evaluation,'' in
  \emph{Advances in Neural Information Processing Systems}, 2014, pp.
  1269--1277.

\bibitem{denil2013predicting}
M.~Denil, B.~Shakibi, L.~Dinh, N.~De~Freitas \emph{et~al.}, ``Predicting
  parameters in deep learning,'' in \emph{Advances in Neural Information
  Processing Systems}, 2013, pp. 2148--2156.

\bibitem{ye2017learning}
S.~Zhe, K.~Zhang, P.~Wang, K.-c. Lee, Z.~Xu, Y.~Qi, and Z.~Ghahramani,
  ``Distributed flexible nonlinear tensor factorization,'' in \emph{Advances in
  Neural Information Processing Systems}, 2016, pp. 928--936.

\bibitem{novikov2015tensorizing}
A.~Novikov, D.~Podoprikhin, A.~Osokin, and D.~P. Vetrov, ``Tensorizing neural
  networks,'' in \emph{Advances in Neural Information Processing Systems},
  2015, pp. 442--450.

\bibitem{monfort2018moments}
M.~Monfort, B.~Zhou, S.~A. Bargal, A.~Andonian, T.~Yan, K.~Ramakrishnan,
  L.~Brown, Q.~Fan, D.~Gutfruend, C.~Vondrick \emph{et~al.}, ``Moments in time
  dataset: one million videos for event understanding,'' \emph{arXiv preprint
  arXiv:1801.03150}, 2018.

\bibitem{liu2009recognizing}
J.~Liu, J.~Luo, and M.~Shah, ``Recognizing realistic actions from videos "in
  the wild",'' in \emph{Proceedings of the IEEE Conference on Computer Vision
  and Pattern Recognition}, 2009, pp. 1996--2003.

\bibitem{srivastava2014dropout}
N.~Srivastava, G.~Hinton, A.~Krizhevsky, I.~Sutskever, and R.~Salakhutdinov,
  ``Dropout: A simple way to prevent neural networks from overfitting,''
  \emph{The Journal of Machine Learning Research}, vol.~15, no.~1, pp.
  1929--1958, 2014.

\end{thebibliography}
